\documentclass[letterpaper, 10 pt, conference]{ieeeconf}  

\IEEEoverridecommandlockouts                              

\usepackage{hyperref}
\usepackage{graphicx}
\usepackage{gensymb}
\usepackage{url}
\usepackage{amsmath} 
\usepackage{amssymb}  
\usepackage{cases}
\usepackage{algorithm}
\usepackage{algpseudocode}
\usepackage{booktabs}
\usepackage{caption}
\usepackage{subcaption}
\usepackage{tikz}
\usetikzlibrary{positioning}

\title{\LARGE \bf
Planning for Aerial Robot Teams for Wide-Area Biometric and Phenotypic Data Collection 
}

\author{Tianshuang Gao$^{1}$, Shashwata Mandal$^{1}$ and Sourabh Bhattacharya$^{1}$
\thanks{*This work was in part supported by grants NSF RI 1816343 and NIFA Award 20176702125965}
\thanks{$^{1}$Department of Computer Science, Iowa State University, Ames, IA 50010, USA
        {\tt\small tsgao@iastate.edu smandal@iastate.edu sbhattac@iastate.edu}}%
}

\begin{document}

\maketitle
\thispagestyle{empty}
\pagestyle{empty}

\begin{abstract}


This work presents an efficient and implementable solution to the problem of joint task allocation and path planning in a multi-UAV platform deployed for biometric data collection in-the-wild. The sensing requirement associated with the task gives rise to an uncanny variant of the traditional vehicle routing problem with coverage/sensing constraints. As is the case in several multi-robot path-planning problems, our problem reduces to an $m$TSP problem. In order to tame the computational challenges associated with the problem, we propose a hierarchical solution that decouples the vehicle routing problem from the target allocation problem. As a tangible solution to the allocation problem, we use a clustering-based technique that incorporates temporal uncertainty in the cardinality and position of the robots. Finally, we implement the proposed techniques on our multi-quadcopter platforms.


\end{abstract}

\section{INTRODUCTION}
In the past decade, there has been a widespread deployment of unmanned aerial vehicles (UAVs) for surveillance in civilian as well as military applications \cite{ollero2007multiple}. A key aspect in such missions is the exploration of interest points \cite{bahceci2003review} for situational awareness. Although UAVs have proven to be highly successfull in wide-area surveillance operations involving identification and tracking of salient entities, their capability is restricted due to the limited amount of onboard power. Hence energy-efficient trajectory planning for these vehicles becomes imperative. This gives rise to several challenges due to the close coupling between the low-level continuous-time optimal control problem involving the dynamics of the individual UAV and the discrete-time combinatorial optimization problems that arise at the team level. In this work, we address a joint allocation and path planning problem that arises when a team of UAVs is deployed to collect biometric/phenotypic data.   



In general, vehicle routing problems reduce to some variant of the famous Travelling Salesman Problem (TSP) \cite{golden1981two, flood1956traveling, laporte1992traveling, gutin2006traveling}. 
In this paper, we deal with a variant of a TSP called as $m$TSP. In $m$TSP, $m$ salesmen are initially located at a depot. Given a set of cities, and a cost metric, the goal is to calculate a set of routes for the $m$ salesmen so that the total sum of the cost of the $m$ routes is minimized. $m$TSP is an NP-hard problem \cite{kiraly2010novel}. However, approximation techniques for TSP (e.g. \cite{golden1981two, flood1956traveling, laporte1992traveling, gutin2006traveling, gorenstein1970printing,bellmore1974transformation} can be used to provide approximate solutions for the $m$TSP since $m$TSP can be reduced to a standard TSP. 

Minimizing the energy consumption of a network to increase its average lifetime is a well-studied problem \cite{rodoplu1999minimum}. In robotic networks, this often translates to solving an optimal control problem for minimizing a metric related to the energy expended by the robot, for example, distance traveled \cite{mohamed2011improved}, time required for task completion \cite{jiang1997time}, wheel rotation \cite{chitsaz2009minimum}, to name a few. In the past, researchers have studied energy-optimal trajectory generation for quadcopters \cite{cowling2010direct} \cite{chamseddine2012flatness} \cite{taamallah2017trajectory} \cite{morbidi2016minimum}. For time-optimal motion planning of quadcopters, a commonly used analytical technique is the Pontryagin Minimum Principle \cite{6094775} \cite{hehn2011quadrocopter} \cite{hehn2012performance}. This approach provides the motion primitives of the optimal trajectory. The challenging part is to use the motion primitives for synthesis of the complete optimal trajectory between an initial and a final state. Non-linear programming has also been utilized \cite{lai2006time} \cite{bouktir2008trajectory} \cite{kahale2014minimum} in time-optimal control. Non-linear programming-based approaches first generate the control points, and then parameterize the generated path in time such that the dynamic constraints are enforced. Moreover, for generating smooth flight path, strategies on minimum snap trajectory planning have been proposed in \cite{mellinger2011minimum} \cite{mellinger2012trajectory} \cite{yu2016optimal}.

The contributions of this paper are as follows: (i) We present a methodology to develop trajectory generation algorithms that can be implemented on aerial vehicles deployed in surveillance applications related to biometric/phenotypic data collection. (ii) We present a clustering approach to the UAV-target allocation problem that is scalable in the number of targets and UAVs (iii) The allocation approach presented in this paper can incorporate temporal uncertainties in the number and position of targets. (iv) The proposed techniques are implementable on a multi-UAV platform as demonstrated by the experiments.

The paper is organized as follows. Section II presents the problem formulation. Section III presents the solution approach. Section IV presents efficient trajectory generation technique for the quadcopter that take into account the sensing requirements and vehicle dynamics. Section V proposes a clustering-based technique for target allocation. Section VI presents the experimental set up and associated results. Section VII presents the conclusion along with the future work.
\vspace{-0.05in}
\section{PROBLEM FORMULATION}
\label{problem_formulation}
We consider a problem in which a team of robots equipped with a vision sensor surveils a region to collect biometric data of targets. The objective of the mobile robots is to acquire a 360$\degree$ view of the target. Since we assume that the robots are fewer in number compared to the targets, each robot has to visit multiple targets. 


Let $T=\{t_1,\dots,t_n\}$ denote the team of $n$ targets and $R=\{r_1,\dots,r_m\}$ denote the team of $m$ robots.  We assume that all the robots start at a point called a {\it depot} denoted as $s$. Let $d_s:T\rightarrow\mathbb{R}$ denote the distance traveled by the robot between the depot and the target. Let $d:T\times T\rightarrow\mathbb{R}$ denote the distance traveled by the robot between targets. A robot $r_i$ is tasked with a sequence $\pi_i$ of $k_i$ targets $K_i$ such that $\bigcup\limits_{i\in R} K_i = T$. Let $\Pi$ be the ordered set of permutations such that $\Pi = \{\pi_i | i \in R\}$. In order to capture a 360$\degree$ scan of the target, the robot rotates a full circle around it before visiting the next target. The radius of the circle is determined by two factors 1) the safe distance from the target 2) distance between the camera and the target required to capture a high quality image. Therefore, the total distance traveled by $r_i$, denoted as $C_i$, to follow a sequence of targets in $\pi_i$ is given by the following expression:

\begin{equation}
C_i(\pi_i) = d_s(\pi_i(1)) + \sum_{j=1}^{k_i-1} d(\pi_i(j),\pi_i(j+1))+ \sum_{j=1}^{k_i} h(\pi_i(j)),\nonumber
\end{equation} 
where $h$ denotes the distance traveled by the robot around a target to acquire a 360$\degree$ scan. 
The total cost incurred by the team of robots associated with an ordered set of permutation $\Pi$ is given as follows:
\begin{equation}
C(\Pi) =  \sum_{i\in R} C_i(\pi_i)
\end{equation}
In order to minimize the energy spent in the surveillance task, we pose the joint allocation and path planning problem for the robots as an optimization problem of minimizing the total distance covered by the team of robots. In other words, the objective is to find the permutation $\Pi^*$ defined as follows:
\begin{equation}\label{prob}
\Pi^*=\arg\min_\Pi C(\Pi)
\end{equation}
In this work, we assume that the robots are quadcopters. For a target modeled as a point and robot modeled as a holonomic vehicle, the problem in (\ref{prob}) is an $m$TSP problem. Our problem is a variant of the $m$TSP problem with additional constraints of target loitering and vehicle dynamics. 


\section{Hierarchical Planning}
\begin{figure}[htpb]
	\centering
	{\includegraphics[width=0.4\textwidth]{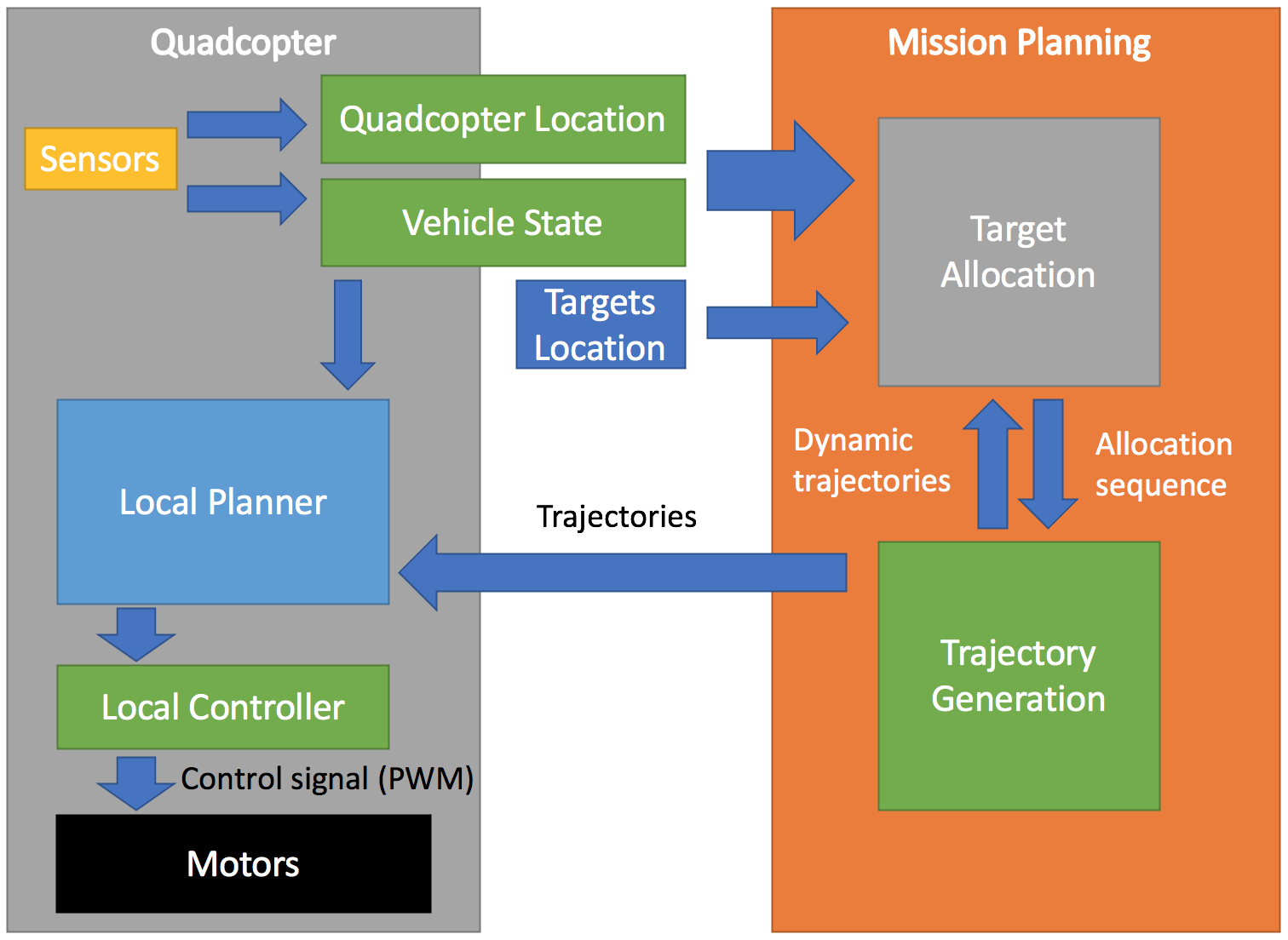}}
	\caption{Planner}
	\label{planner}
\end{figure}
Figure \ref{planner} shows our proposed solution. The overall planner can be divided into two major modules:
\begin{enumerate}
    \item Mission Planning: The role of this module is to generate the route along with the supported trajectory for the system. The input to the module is the current location of the target. It computes the final trajectory and forwards it to the planner module which is located on-board the quadcopters. This is a part of the ground station.
    \item Quadcopter : The role of this module is to serve as the actuator and is a part of every quadcopter.
    \begin{enumerate}
        \item Sensors : They are responsible for keeping track of the current system states
        \item Quadcopter Location : This refers to the GPS module on-board which keeps track of the physical location of the quadcopter
    \end{enumerate}
\end{enumerate}

The overall system works as follows: The input to the Mission planning module is the state of the quadcopter and location of the targets. The target allocation module processes the incoming data and generates the allocation and visiting sequence for each quadcopter periodically. Trajectory generation module receives the output from the target allocation module. It generates the trajectories based on the dynamic model of the quadcopter. The trajectory is fed back to the target allocation module in order to re-allocate the targets for the incoming/moving targets in a dynamic scenario. The quadcopter takes the generated trajectory as its mission command. The local planner combines the trajectory and the current state of the quadcopter in order to compute the desired motion. Finally, the local controller sends the control signal (Pulse-width modulation) directly to the motors to follow the trajectory and complete the task. The hierarchical nature of the planner stems from the two-step trajectory generation strategy with the first stage being path generation followed by its upgradation to a minimum time trajectory at the end of the two stages.

\vspace{-0.05in}
\section{Trajectory generation}
\label{}

 In this section, we describe the different phases of the quadcopter's proposed trajectory.
\subsection{Loitering Phase}
In the loitering phase, the quadcopter scans a target while traversing a circular trajectory of radius $r$ at its maximum speed at a constant height $h$. Figure \ref{frames} shows the free-body diagram of a quadcopter traversing a circular trajectory around the target (the human) in the inertial frame $\mathcal{W}$ and the body frame $\mathcal{B}$. Equating the forces along the tangential, radial and vertical directions for leads to the following force balance equations: 

\begin{figure}[thpb]
	\centering
	{\includegraphics[width=0.45\textwidth]{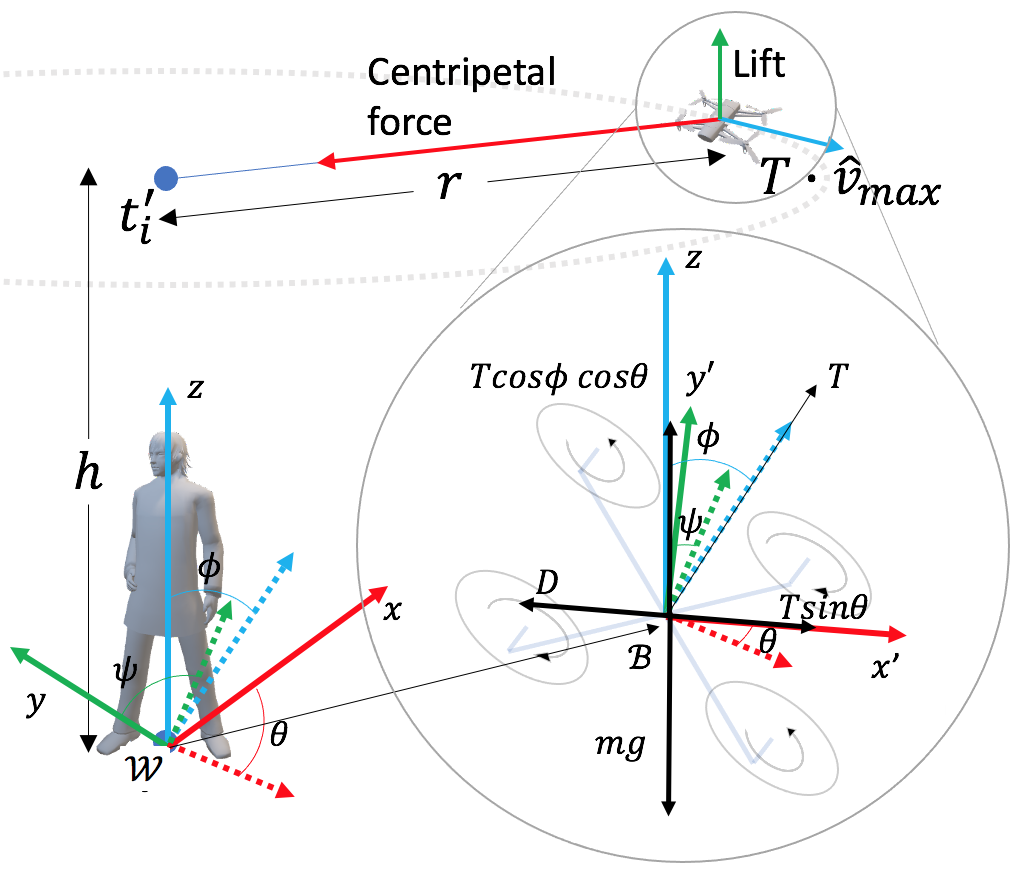}}
	\caption{The figure shows the free body diagram of a quadcopter moving in a circular trajectory around a target. The inset shows the configuration variables associated with the quadcopter in the body coordinate system.}
	\vspace{-0.1in}
	\label{frames}
\end{figure}
\vspace{-0.1in}

\begin{eqnarray}
	\boldsymbol{T} \cos{\phi} \cos{\theta}& = &m \boldsymbol{g}\label{eq1}\\ \frac{m \boldsymbol{v}_{max}^2}{r} &=& \boldsymbol{T} \cos{\theta} \sin{\phi}\label{eq2}\\\boldsymbol{T} \sin{\theta} &=& \boldsymbol{D}, \label{eq3}
\end{eqnarray}
where $\boldsymbol{e} = 
\begin{bmatrix}
\phi & \theta & \psi 
\end{bmatrix}^T
\in \mathbb{R}^3$ denotes the Euler angles, $m$ is the mass of the quadcopter, $\boldsymbol{v}_{\max}$ denote the maximum speed of the quadcopter, $\boldsymbol{T}$ denotes the thrust generated in the direction of the motors, $\boldsymbol{D} (= -\frac{1}{2} \ c_d \ \boldsymbol{v}^2 \cdot \hat{\boldsymbol{v}})$ denotes the air-drag of the quadcopter, $c_d$ is the air drag coefficient that depends on area of cross section and air density \cite{batchelor2000introduction}. Eliminating $\theta$ and $\phi$ in (\ref{eq1}), (\ref{eq2}) and (\ref{eq3}) leads to the following equation for $\boldsymbol{v}_{max}$ in terms of the known parameters:
\begin{equation}
\label{eqn_v_max}
\boldsymbol{v}_{max}=\left[\frac{T^2-m^2g^2}{\frac{m^2}{r^2}+\frac{c^2_d}{4}}\right]^{\frac{1}{4}}
\end{equation}

\subsection{Transition Phase}
\label{approaching_strategy}
\begin{figure}[thpb]
	\centering
	{\includegraphics[width=0.4\textwidth]{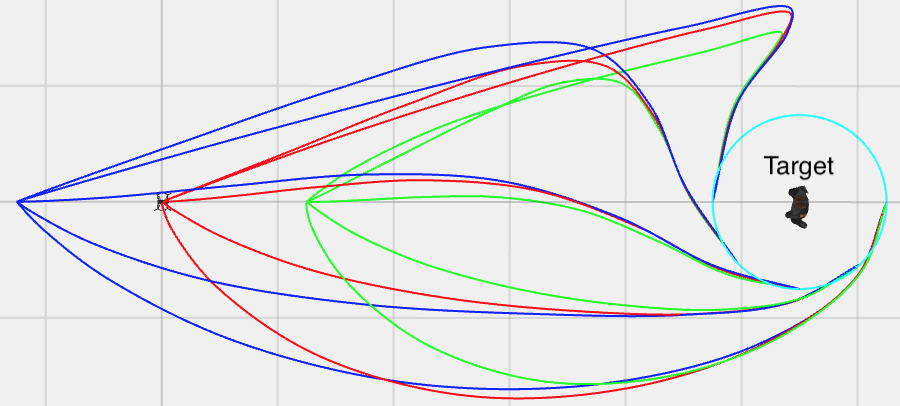}}
	\caption{Minimum-time trajectories for three different initial states of the quadcopter for five equi-spaced terminal points on the semicircle. The target circle is shown in light blue color.}
	\label{fig_move_to_a_target}
\end{figure}
\vspace{-0.1in}

In this phase, the quadcopter leaves the circle around one target and arrives on the circle around the next target. The state it leave a circle is where it entered the circle earlier. The minimum-time control problem of the quadcopter can be formulated as the following Bolza problem (\cite{}):
\begin{equation}\label{cost}
J=\int^{t_f}_{0}1\quad dt,    
\end{equation}
where $t_f$ denotes the time at which the quadcopter reaches the circle. Due to the non-linearity in the dynamics of the quadcopter, (\ref{cost}) is solved using Optimtraj library \cite{kelly}. Since the non-linear optimization is a two-point boundary value problem, the final circle is discretized to determine the values of the state to enter the circle. At the final state, the direction of rotation of the quadcopter around the target can be clockwise or anti-clockwise. This leads to two optimization problems that need to be solved for a given final state. Fig.~\ref{fig_move_to_a_target} shows the minimum-time trajectory for three different initial distance from the circle for five equispaced terminal points on the semicircle. 

\begin{algorithm}
	\algrenewcommand{\algorithmiccomment}[1]{\hskip1em$/*$ #1 $*/$}
	\renewcommand{\algorithmicrequire}{\textbf{Input:}}
	\renewcommand{\algorithmicensure}{\textbf{Output:}}
	\caption{Trajectory Generation from Target $s$ to Target $t$}
	\label{traj_alg}
	\begin{algorithmic}[1]
		\Require{Point $p$ where a quadcopter leaves target $s$, Target $t$, radius $r$ }
		\Ensure{Trajectory $T$}
		\Function{TrajectoryGenerator}{$p, t$}
		\State Discretize points on circles $t$
		\State Sample entry points $S$ on circle $t$
		\For{each entry points $e \in S$}
		    \State Calculate the time-optimal trajectory between $p$ and $e$ using Non-linear optimizer
	    \EndFor
	    \State Find the trajectory $T$ that has the shortest time
		\State \Return $T$
		\EndFunction
	\end{algorithmic}
\end{algorithm}
\vspace{-0.1in}
\section{Clustering-based Target Allocation}
\label{allocation}
		

In this section, we address the allocation problem for the team of quadcopters. Assuming the quadcopters to be salesmen and the targets to be cities, the original problem as posed in (\ref{prob}) is an $m$TSP problem which is well-known to be NP-hard \cite{kiraly2010novel}. Past efforts in providing a solution to the $m$TSP problem involve heuristic \cite{junjiedingwei} and approximation \cite{frederickson}. Since the space in our problem is metrizable, we can reduce it to a standard $\Delta{\text -}TSP$ problem. In $\Delta{\text -}TSP$, there are $n$ cities with a distance function in the form of $d:[n]\times[n]\rightarrow\mathbb{R}$ where $d$ is set in a metric space. The objective is to find a permutation $\pi$ which minimizes the total distance $D = \sum_{j=1}^{n-1} d(\pi_i(j),\pi_i(j+1) + d(\pi(1), \pi(n))$. Therefore, problem (\ref{prob}) is equivalent to solving a multi-variate version of the standard $\Delta{\text -}TSP$.

Even though $\Delta{\text -}TSP$ is known to be NP-Hard \cite{cormen}, the Christofides' algorithm provides a $\frac{3}{2}$ approximation algorithm for $\Delta{\text -}TSP$ \cite{christofides1976worst}. Leveraging on Christofides' algorithm, Algorithm 2 presents a hierarchical approach to the allocation problem using a clustering technique. The crucial part of the solution is to first partition the cities into $m$ clusters using a $K$-means clustering algorithm \cite{nallusamy2010optimization} (where $m$ is the number of quadcopters) and then compute a tour for each of these $m$ clusters using Christofides algorithm. In Line 2, the targets are divided into $m$ clusters. It may be noted that the $K$-means algorithm may be replaced by any other clustering algorithm. An advantage of this approach is its flexibility in accommodating additional targets in the environment without recomputing the clusters. Line 3 is an optional optimization step for decreasing the number of targets for surveillance. The central idea here is to use hierarchical clustering \cite{Defays1977AnEA} to represent targets which are really close to each other as a single target. In Lines 5-8, we run Christofides Algorithm on each of the above mentioned clusters. Finally, once the tours are constructed for each of the clusters, the tours are forwarded to the trajectory generation algorithm.

\begin{algorithm}
	\algrenewcommand{\algorithmiccomment}[1]{\hskip1em$/*$ #1 $*/$}
	\renewcommand{\algorithmicrequire}{\textbf{Input:}}
	\renewcommand{\algorithmicensure}{\textbf{Output:}}
	\caption{Assigning $n$ targets to $m$ quadcopters}
	\label{target_assignment}
	\begin{algorithmic}[1]
		\Require{$m$ quadcopters $Q$, depot location $s$, $n$ starting targets along with their locations $T$, distance functions $d_s$ and $d$}
		\Ensure{Targets assignment $A$ - $m$ ordered sets with each set representing the order of targets visited by that quadcopter}
		\Function{TargetAssignment}{$T, Q, s, d, d_s$}
		\State $A_{K-means} \gets$ \Call{K-Means}{$m, T$}  \Comment{Call K-means and store assignment in $A_{K-means}$}
		\State $A'_{K-means} \gets$ \Call{Trucate-Targets}{$A_{K-means}$}
		\For{each set $q \in T'_{K-means}$}
			\State Construct graph $G_q$ from $T$ using $q$ and $d$
			\State $A_q \gets$ \Call{Christofides}{$G_q$} \Comment{$A_q$ contains the ordered set of targets visited in the TSP tour}
			\State Find the closest point in $A_q$ from $s$ and set it as the beginning of the tour
			\State Append $A_q$ to $A$
		\EndFor
		\State \Return $A$
		\EndFunction
	\end{algorithmic}
\end{algorithm}



\begin{figure}[thpb]
	\centering
	{\includegraphics[width=0.35\textwidth]{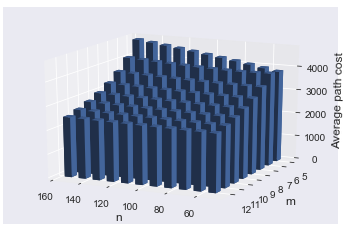}}
	\caption{Figure shows the variation of the average path cost per quadcopter for different values of $n$ and $m$. For each $m$ and $n$, the average is computed over 10,000 simulations )}
	\label{barchart}
\end{figure}

\begin{figure}[thpb]
	\centering
	{\includegraphics[width=0.4\textwidth]{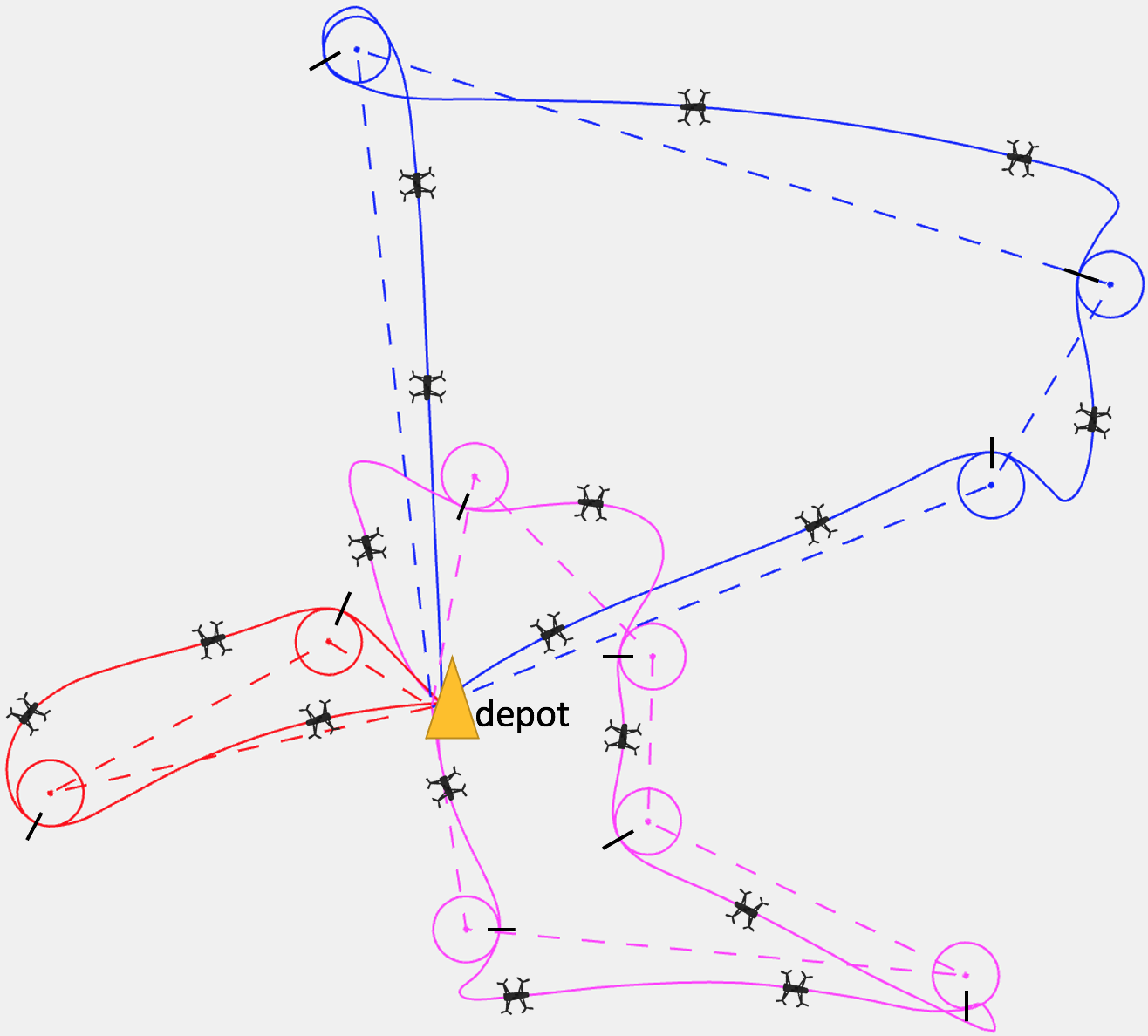}}
	\caption{Figure shows the trajectory (solid lines) of the quadcopters based on the numerical optical control proposed in Section IV, and the (dashed lines) trajectory generated by Algorithm 2.}
	\label{actualvsgen}
\end{figure}

Next, we describe our simulation results. Figure \ref{actualvsgen} shows the output of Algorithm 2 for 10 targets and 3 quadcopters. The targets allocated to a specific quadcopter have the same color. The dotted lines denote the path of the quadcopters in each cluster. Figure \ref{barchart} shows the average cost (path length) per quadcopter for different values of $m$ (number of quadcopters) and $n$ (number of targets). The simulation is performed by choosing $n$ samples as target locations from a uniform distribution over a square region of $X:[-1000, 1000]\times Y:[-1000, 1000]$. Algorithm \ref{target_assignment} is run 1000 times for each value of $n$ and $m$. We can see that the path cost per quadcopter decreases as the number of quadcopters increases which is intuitive. Since the problem considered in this work deals with quadcopters that acquire a 360$\degree$ scan of the target, we need to incorporate sensing constraints and robot dynamics into the path planning algorithm. The solid lines in Figure \ref{actualvsgen} shows the trajectory of the quadcopters based on implementing Algorithm~\ref{traj_alg} on the allocation obtained from Algorithm 2.

\begin{figure}[htpb]
	\centering
	{\includegraphics[width=0.35\textwidth]{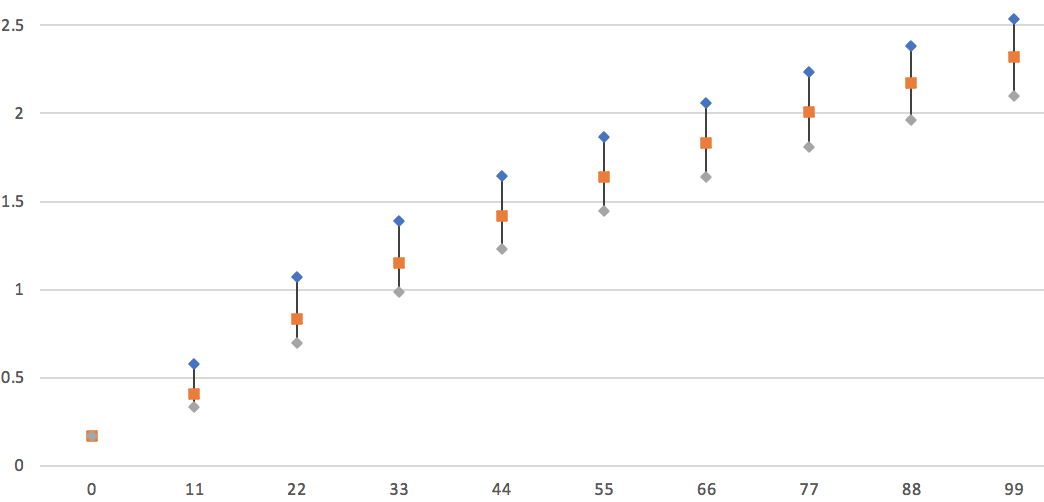}}
	\caption{Avg-Max-Min time radio between holonomic vehicle without motion constraint and our result.}
	\label{holonomic}
\end{figure}

Since the trajectories of the quadcopter are obtained from numerical optimal control, it is challenging to obtain a theoretical gap between the cost incurred by the team when the quadcopters are approximated by a holonomic vehicle. Figure \ref{holonomic} shows the average, maximum and minimum ratio between the minimum time required by a holonomic vehicle and a quadcopter for the transition phase for different initial position of the vehicles. From the simulation results, we can conclude that the distance traveled by quadcopters is at most 3 times that of a holonomic vehicle. For an $m$TSP problem, \cite{frederickson} shows that if the number of paths is $m$ then the overall approximation factor can be further reduced to $\frac{3}{2}-\frac{1}{m}$. Combining the $\frac{3}{2}-\frac{1}{m}$ approximate bound of christofides with the empirically obtained bound of 3 from simulation results leads to an overall $4.5-\frac{3}{m}$ approximation factor for the cost incurred by the quadcopters relative to the optimal route for holonomic vehicles using the allocation technique proposed in Algorithm 2.
\vspace{-0.1in}


\subsection{Incorporating Variable and Mobile Targets}

Since the overall motivation is to build a deployable multi-quadcopter platform for real-time surveillance, the system should be able to adapt to uncertainties in the number and location of the targets. The technique proposed in the previous section for static targets can be considered as a snapshot of the dynamic scenario regarding the state of the targets. In this section, we extend the static algorithm to address situations in which new targets may appear or old targets may disappear from the surveillance region. 

In Algorithm 2, the quadcopters return to the depot after completing their original tours. Algorithm \ref{dynamic_target_assignment} doesn't allow the quadcopters to return to the depot but instead initiates a new tour (Line 7-9) of the updated cluster (Line 4). The algorithm includes a target absorption phase where a new target is absorbed by the existing clusters based on their distance from their cluster centers (Line 4). An additional step includes re-evaluating the TSP circuit for each cluster once the previous iteration of drone surveillance is finished(Line 7-9). 

\begin{algorithm}
	\algrenewcommand{\algorithmiccomment}[1]{\hskip1em$/*$ #1 $*/$}
	\renewcommand{\algorithmicrequire}{\textbf{Input:}}
	\renewcommand{\algorithmicensure}{\textbf{Output:}}
	\caption{Assigning targets to $n$ quadcopters and execute trips}
	\label{dynamic_target_assignment}
	\begin{algorithmic}[1]
		\Require{$m$ quadcopters $Q$, depot location $s$, $n$ targets along with their locations $T$, distance functions $d_s$ and $d$}
		\Function{Dynamic Assignment}{$T, Q, s, d, d_s$}
		\State $A_{K-means} \gets$ \Call{K-Means}{$M$}  \Comment{Call K-means and store assignment in $A_{K-means}$}
		\State Move all quadcopters to the nearest targets in their clusters
		\State Asynchronously assign any incoming targets to their nearest cluster 
		\State $A'_{K-means} \gets$ \Call{Trucate-Targets}{$A_{K-means}$}
		\For{each set $q \in A'_{K-means}$} \Comment{Run each these in parallel}
			\State Construct graph $G_q$ from $T$ using $q$ and $d$
			\State $A_q \gets$ \Call{Christofides}{$G_q$} \Comment{$A_q$ contains the ordered set of targets visited in the TSP tour}
			\State Find the closest point in $A_q$ from $s$ and set it as the beginning of the tour if this is the first iteration
			\State Call \Call{Trip}{$A_q$}
		\EndFor
		\EndFunction
	\end{algorithmic}
\end{algorithm}

Given a scenario where a set of $n$ original targets exist and $r$ new targets are introduced, there are two ways of solving the surveillance problem. The first method is to use Algorithm \ref{target_assignment} to finish surveillance of the original targets. After the new targets are introduced, complete a fresh execution of Algorithm \ref{target_assignment} on the updated set of targets. The second method is to use Algorithm \ref{dynamic_target_assignment} to complete surveillance of the original targets. After the new targets are introduced, they are absorbed by the existing clusters, and Algorithm \ref{dynamic_target_assignment} completes surveillance of the new clusters in its second iteration. 

\begin{figure}[ht] 
  \begin{subfigure}[b]{0.5\linewidth}
    \centering
    \includegraphics[width=\linewidth]{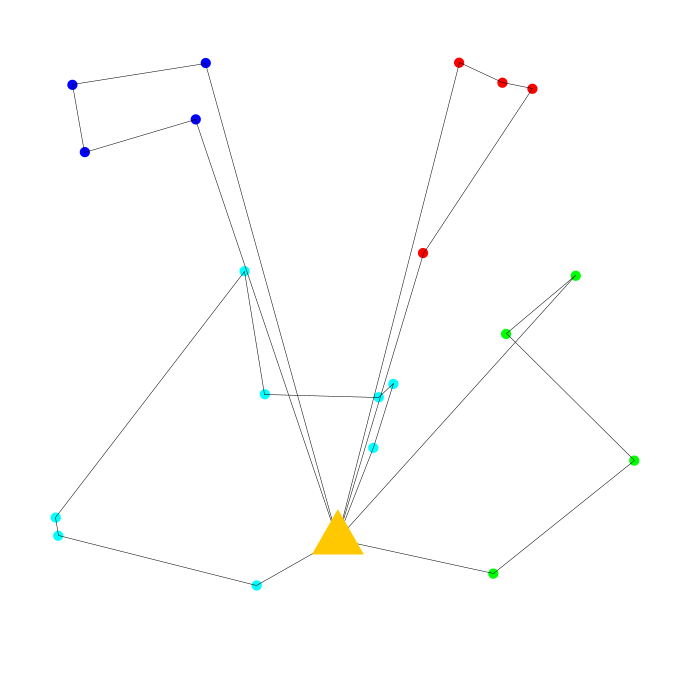} 
    \caption{Algorithm \ref{target_assignment} - Run 1} 
    \label{techniques:11} 
    \vspace{1ex}
  \end{subfigure}
  \begin{subfigure}[b]{0.5\linewidth}
    \centering
    \includegraphics[width=\linewidth]{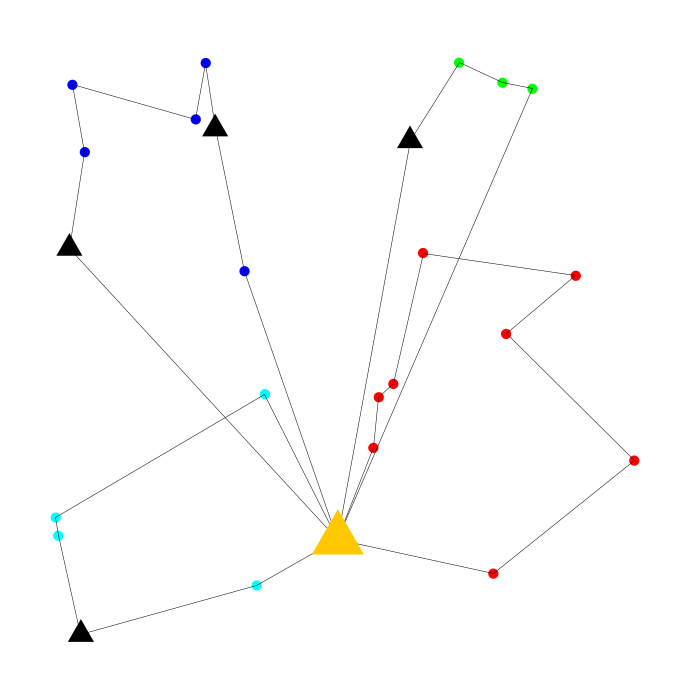} 
    \caption{Algorithm \ref{target_assignment} - Run 2 with new targets} 
    \label{techniques:12} 
    \vspace{1ex}
  \end{subfigure} 
  \begin{subfigure}[b]{0.5\linewidth}
    \centering
    \includegraphics[width=\linewidth]{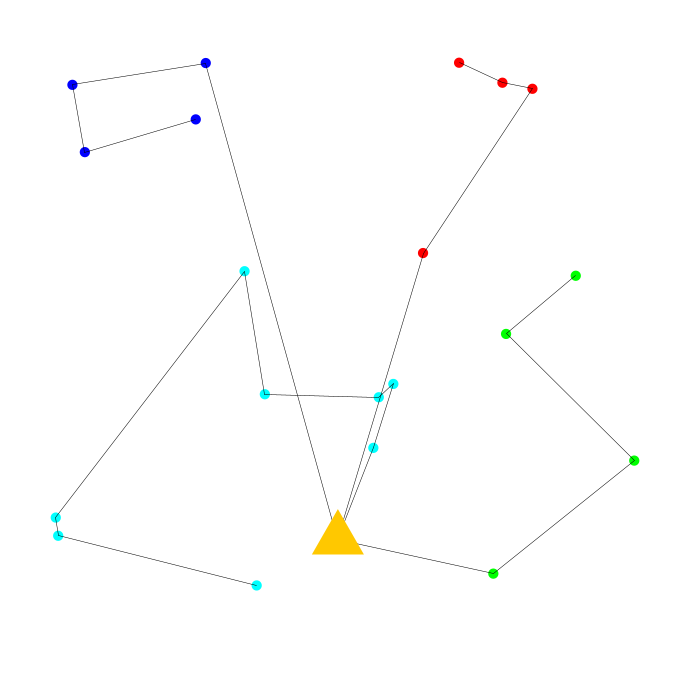} 
    \caption{Algorithm \ref{dynamic_target_assignment} - Iteration 1} 
    \label{techniques:21} 
  \end{subfigure}
    \begin{subfigure}[b]{0.5\linewidth}
    \centering
    \includegraphics[width=\linewidth]{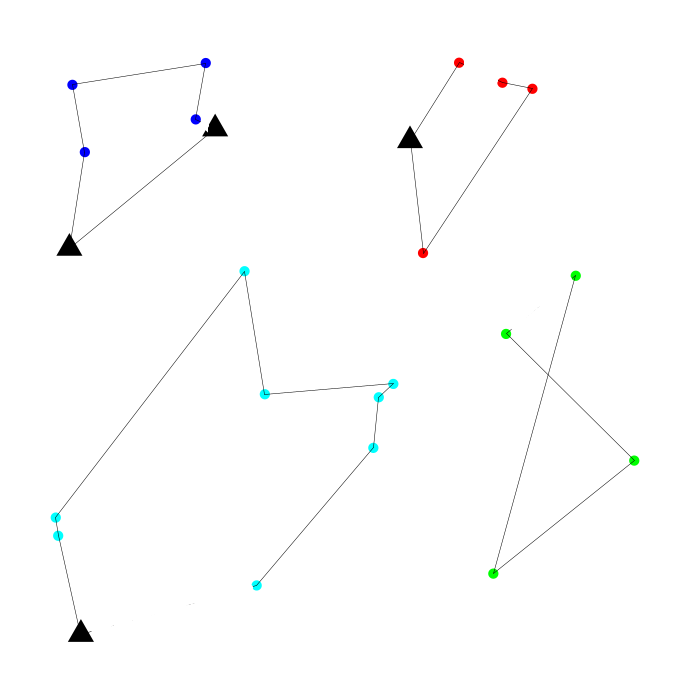} 
    \caption{Algorithm \ref{dynamic_target_assignment} - Iteration 2 with new targets} 
    \label{techniques:22} 
  \end{subfigure}
  \caption{Figure shows the two procedures applied to the dynamic scenario where new targets (represented by black triangles) arrive. The orange triangle represents the depot.}
  \label{techniques} 
\end{figure}

Figure \ref{techniques} demonstrates the procedures when new targets are introduced in the environment. We can see that Procedure 1 relies on completing the tour and returning to the depot where as Procedure 2 completes the tour and stays on the final target waiting for the new iteration. This creates a difference in terms of the total path cost.

\begin{figure}[ht] 
  \begin{subfigure}[b]{0.5\linewidth}
    \centering
    \includegraphics[width=\linewidth]{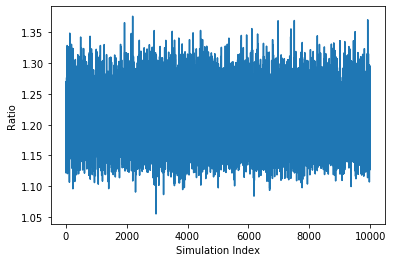} 
    \caption{$r = 32$} 
    \label{ratio_sim_r32} 
    \vspace{1ex}
  \end{subfigure}
  \begin{subfigure}[b]{0.5\linewidth}
    \centering
    \includegraphics[width=\linewidth]{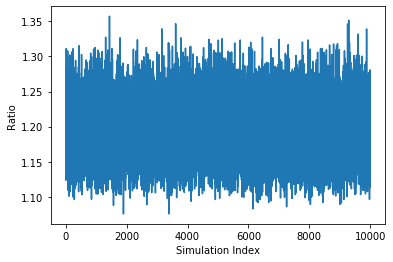} 
    \caption{$r = 64$} 
    \label{ratio_sim_r64} 
    \vspace{1ex}
  \end{subfigure}
 \begin{subfigure}[b]{0.5\linewidth}
    \centering
    \includegraphics[width=\linewidth]{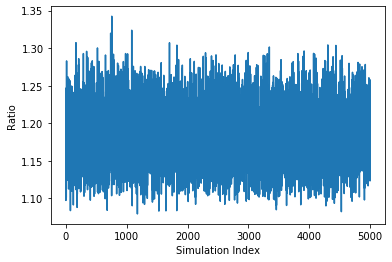} 
    \caption{$r = 100$} 
    \label{ratio_sim_r100} 
  \end{subfigure}
  \begin{subfigure}[b]{0.5\linewidth}
    \centering
    \includegraphics[width=\linewidth]{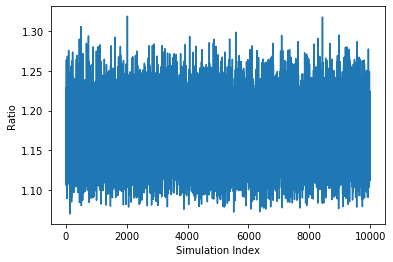} 
    \caption{$r = 150$} 
    \label{ratio_sim_r150} 
  \end{subfigure}
  \caption{Comparing the ratio ($\rho$) of the path lengths in Procedure 1 and Procedure 2, where $n = 100$, $m = 8$ and number of simulations = $30000$}
  \vspace{-0.1in}
  \label{ratio_sim} 
\end{figure}

Figure \ref{ratio_sim} shows the results for the simulation where new targets are introduced after the old targets have been processed. The simulation is initiated with $100$ targets ($n$) and $8$ quadcopters ($m$). Once the first round of processing is completed, $r$ new targets are introduced. After the second round of processing, the path lengths generated from the two procedures are compared and analysed. The ratio ($\rho = \frac{Path Length_{Procedure1}}{Path Length_{Procedure2}}$) is the metric for the comparison. We see the results for the above mentioned simulation for $r = 32,64,150$. We see that the path lengths for Procedure 2 is less than that for Procedure 1 even after 150 new targets are introduced. This clearly demonstrates how Algorithm \ref{dynamic_target_assignment} outperforms Algorithm \ref{target_assignment} at handling dynamic scenarios.


\section{EXPERIMENTs}
\label{hardware}
The strategy proposed in this paper is implemented on a multi-UAV testbed developed in our lab called {\it Cy-Eye} \cite{gao2018novel}. 

\subsection{Cy-Eye Architecture}
\begin{figure}[htpb]
	\centering
	{\includegraphics[width=0.45\textwidth]{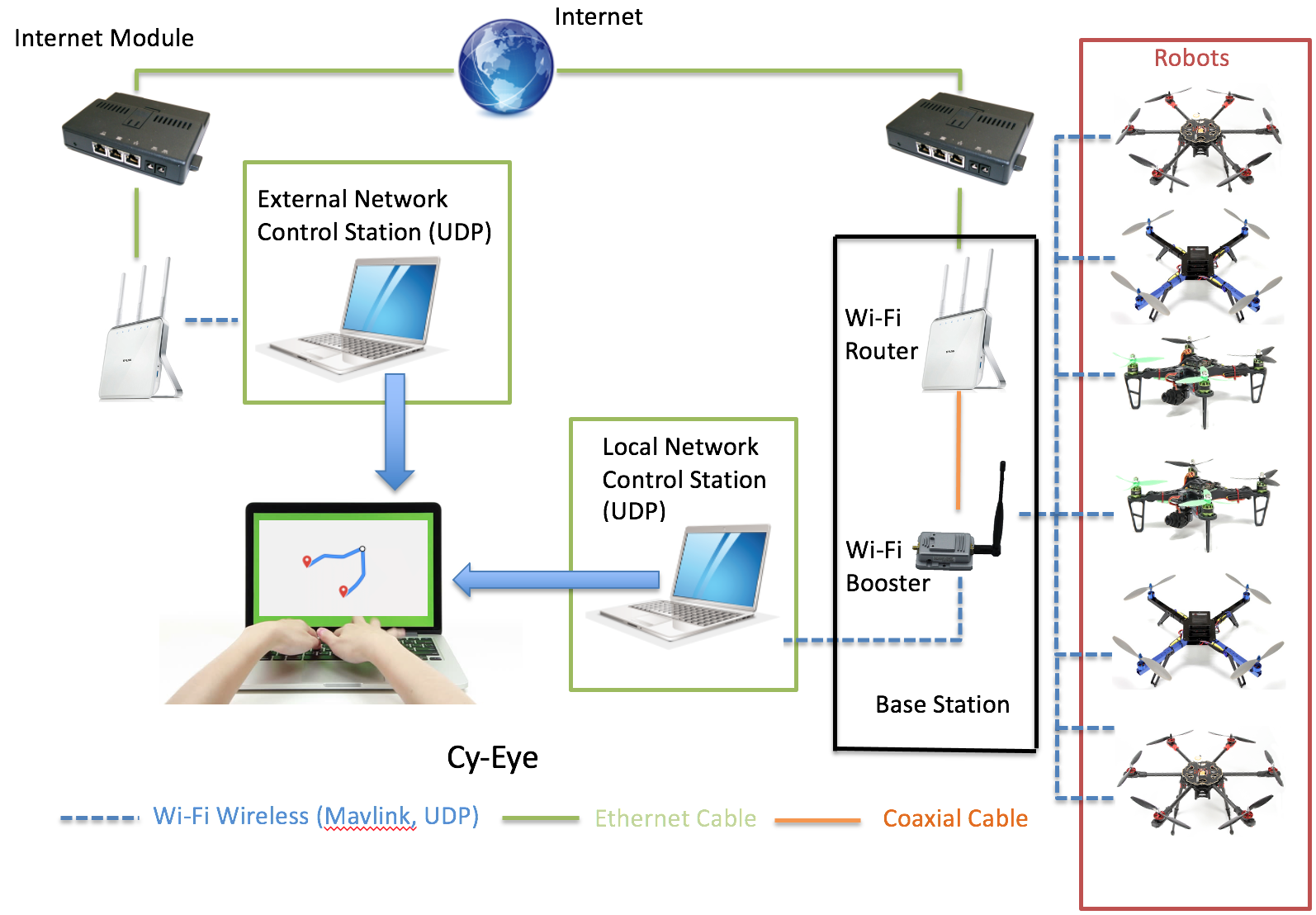}}
	\caption{Communication architecture of Cy-Eye.}
	\label{system}
\end{figure}

Figure \ref{system} shows the communication architecture. Two computer boards, Pixhawk (STM32 chip) and Nvidia Jetson Nano are carried by each quadcopter. The ground control is a laptop running APMPlanner. Ardupilot \cite{ardupilot} is an open source autopilot control running on Pixhawk. It processes sensor data and calculates the desired feedback signal to determine the input to the motors. The autopilot is fully programmable and can have a companion on-board computer (Nvidia Jetson Nano) linked through UART serial protocol. A companion computer is used as a communication bridge between the autopilot and our ground control, and also makes online decision possible during the flight. The GPU on the companion on-board computer allows on-board image processing. MAVProxy, run on the companion computer, provides a control platform that links the autopilot and ground control.

We deployed the quadcopters to encircle 8 targets in an open space (50x50m). The ground control allocates the targets for each quadcopter by using Algorithm~\ref{target_assignment}. The trajectories generated by Algorithm~\ref{traj_alg} are sent to each quadcopter through UDP/MAVLink protocol. The parameters in the dynamic model (Inertial, propulsion, air density, thrust, etc.) are measured based on our custom-build quadcopter (450mm). In the experiment, we set the flying altitude between 1m and 2m. Once the trajectories are received, the quadcopters start their task automatically by following the trajectories. 

\begin{figure}[htbp]
	\centering
	{\includegraphics[width=0.4\textwidth]{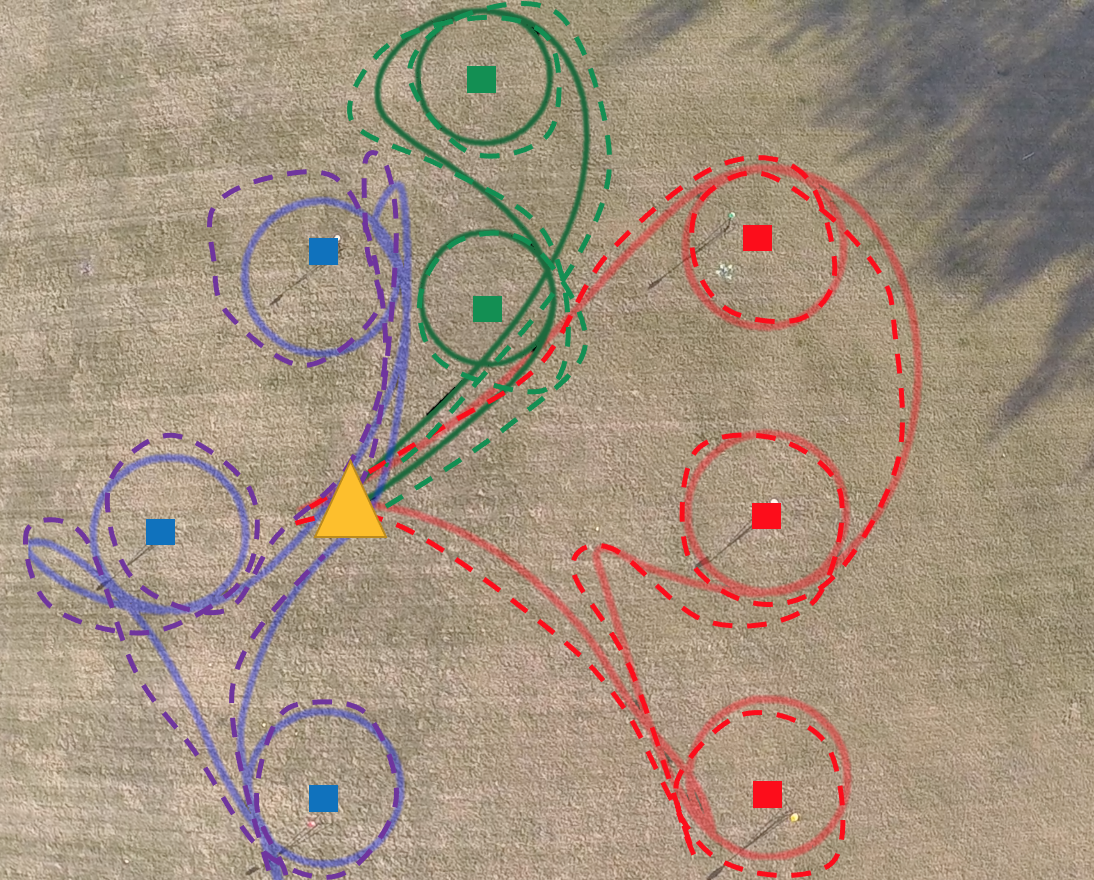}}
	\caption{Figure shows the paths generated from Algorithm~\ref{traj_alg} (solid), and the actual paths (dashed) traversed by the quadcopters during the experiment.}
	\label{setup}
\end{figure}

Figure \ref{setup} shows the path traversed by three quadcopters using our proposed algorithm. The actual flight paths (shown with dashed line) are longer than the trajectories generated by Algorithm 1. Although a longer path leads to a longer flying time, the result shows that the system could successfully complete the task of encircling the targets. A video accompanying the submission provides details regarding the experiments.

\section{Conclusion and future work}
In this work, we present a strategy to generate trajectories that can be implemented on aerial robots deployed in a surveillance application. The problem is well-known to be NP-hard. The hierarchical approach that we proposed divides this problem into 2 sub-problems (Targets allocation and trajectories generation). We present a clustering approach to the allocation problem that is scalable in the number of targets and robots. A non-linear programming-based trajectories generation approach is presented. We investigate the proposed strategies through extensive simulation. Finally, the same strategy was applied on a multi-UAV platform in the outdoor environment. 

As a future research direction, we plan to investigate the performance of the system for several applications. Of particular interest, are applications related to intruder identification and livestock phenotyping. An ongoing research direction is the detection of face masks for preventing respiratory infections diseases in humans. We are specifically interested in implementing real-time machine learning algorithms in the quadcopter perception loop.  

\newpage
\bibliographystyle{IEEEtran}
\bibliography{ref}
\end{document}